# EDocNet: Efficient Datasheet Layout Analysis Based on Focus and Global Knowledge Distillation

Hong Cai Chen, Longchang Wu, Yang Zhang

***Abstract*—When designing circuits, engineers obtain the information of electronic devices by browsing a large number of documents, which is low efficiency and heavy workload. The use of artificial intelligence technology to automatically parse documents can greatly improve the efficiency of engineers. However, the current document layout analysis model is aimed at various types of documents and is not suitable for electronic device documents. This paper proposes to use EDocNet to realize the document layout analysis function for document analysis, and use the electronic device document data set created by myself for training. The training method adopts the focus and global knowledge distillation method, and a model suitable for electronic device documents is obtained, which can divide the contents of electronic device documents into 21 categories. It has better average accuracy and average recall rate. It also greatly improves the speed of model checking.**

## I. Introduction

Electronic component documentation is essential material for circuit design. Engineers need to extract information from a vast amount of component documentation when designing circuits, which is time-consuming. Using a machine learning model to parse electronic component documentation is a new approach [1-3]. However, due to the complexity and unique structures in electronic component documentation, traditional document analysis is not accurate as expected.

Parsing electronic component documentation [4] is complex due to the vast number of components, and their varying attributes like shape, size, color, and markings. The documents are packed with model numbers, parameters, specs, and functions, often presented in various formats including text, symbols, and tables. Accurately extracting and interpreting this information remains a key issue. The complexity is further heightened by the frequent use of diagrams, symbols, and tables, which are non-textual and contribute to the intricate layout of these documents.

The primary methods employ detection models to classify layout content. Examples of such models include Faster R-CNN [5], Mask R-CNN [6], and YOLOv8 [7]. In recent years, with the rapid development of large models, significant progress has been made in document layout analysis using large models, such as DiT [8], LayoutLM [9-11], etc. However, the existing document layout analysis models are primarily designed for general documents and still face certain limitations when it comes to analyzing documents related to electronic components. These limitations stem from a lack of an insufficient ability to recognize the complex layouts found in electronic component documentation. Therefore, despite the significant advancements in the field of layout analysis, further research and optimization are needed to deal with specialized document types like electronic component documents. Additionally, these models mainly categorize documents into five types [12]: text, title, table, figure, and list. For the analysis of electronic component documents, the various diagrams within the document [13] play a crucial role in understanding the components. Therefore, it is essential to further subdivide the documents into subdivision types. Moreover, the captions of diagrams and tables also play a guiding role in the analysis of diagrams and tables, but current models often categorize them as text parts and cannot links them with the corresponding diagrams.

Despite these challenges, accurately parsing electronic component documentation is crucial [14, 15]. It can greatly improve the speed of information retrieval [16], reduce the costs of manually identifying electronic components, and boost the efficiency of design processes [17]. Therefore, this paper conducts a thorough analysis of electronic component documents and develop a specialized model that captures their unique features, thereby improving the efficacy of parsing capacity for such documents. This paper aims to achieve the following objectives:

- This paper proposes the EDocNet model, which is designed specifically for the layout analysis of electronic device documents. This model utilizes a focal and global knowledge distillation training method, enabling it to effectively handle the refinement of specific target categories within such documents.
- Through comprehensive experiments, it showcases the excellent performance of the EDocNet model. It outperforms traditional object detection models like Faster R-CNN, YOLOv8 and large models such as LayoutLMv3 and DiT in terms of average precision, average recall, and significantly reduces the training and prediction time per image. This enhanced efficiency makes it highly suitable for analyzing multi-page electronic device documents.

The rest of the paper is organized as follows: Section II summaries the current research on document analysis. The

This paragraph of the first footnote will contain the date on which you submitted your paper for review. It will also contain support information, including sponsor and financial support acknowledgment. For example, "This work was supported in part by the U.S. Department of Commerce under Grant BS123456."

S. B. Author, Jr., was with Rice University, Houston, TX 77005 USA. He is now with the Department of Physics, Colorado State University, Fort Collins, CO 80523 USA (e-mail: author@lamar.colostate.edu).

detailed formats of electronic component document are presented in Section III. Then, the proposed method is demonstrated in Section IV. Section V shows the performance of the proposed method with the evaluations on electronic component documents and comparisons with state-of-art document analysis tools. Finally, the conclusion is drawn in Section VI.

## II. RELATED WORK

In the field of document layout analysis, recent works mainly focus on dataset construction and deep learning.

### *A. Document Datasets*

In the field of document layout analysis, several large-scale manually annotated datasets have been established for model training and evaluation. The DocLayNet dataset, introduced by Pfitzmann *et al.* [18], comprises a diverse collection of 80,863 pages from various sources, including financial reports, scientific articles, legal regulations, government tenders, manuals, and patents, offering detailed bounding box annotations for 11 distinct layout elements. The PubLayNet dataset, created by Zhong *et al.* [19], encompasses over 360,000 pages with annotations for typical document layout elements such as text, headings, lists, figures, and tables, generated by matching PDF and XML formats from the PubMed Central Open Access subset. Li *et al.* [20] presented the DocBank dataset, which contains 500,000 pages of documents with fine-grained token-level annotations, constructed from a vast array of LaTeX-compiled PDF files and annotated using weak supervision methods for efficient and high-quality annotation acquisition. Cheng *et al.* [21] contributed the M6Doc dataset, comprising 9,080 modern document images across seven subsets, including scientific articles, textbooks, exam papers, magazines, newspapers, notes, and books, in three formats: PDF, photographed documents, and scanned documents, totaling 237,116 annotated instances. These datasets have significantly advanced the field by covering a multitude of document types and layouts. However, they are predominantly geared towards general documents. The representation of specific document types, such as datasheets and electronic design documents, is limited within these datasets, rendering them less suitable for our work, which necessitates more specialized datasets for effective model training and evaluation in these niche areas.

### *B. Document Analysis Methods*

With the rapid development of deep learning technology, significant progress has been made in document layout analysis. It can be categorized into two types of methods: target detection models and large models. Target detection models view document analysis as a unique object recognition challenge and employ standard, readily available detectors. The fast CNN method proposed by Oliveira and Viana [22] and the Transformer-based approach by Yang and Hsu [23], has significantly enhanced the performance of analysis. Faster R-CNN by Ren *et al.* [5] introduced region proposal networks, improving real-time object detection, while DETR by Carion *et al.* [24] employed Transformers to eliminate complex post-processing steps, improving global feature capture. Object detection is a rapid growing field. Yi *et al.* [25] and Grüning *et al.* [26] has improved document image processing accuracy, while Shi *et al.* [27] enhanced page object detection through a lateral feature enhancement network. The research of by Saha *et al.* [28] on graphical object detection and Zhang *et al.* [29] on exploration of document layout analysis methodologies have contributed valuable perspectives to this field. Meanwhile, Jocher *et al.* [30] further advanced real-time detection with the Ultralytics YOLO model, and DINO model [31] refined DETR with denoising anchor boxes, improving object detection accuracy and stability. However, although a variety of models have been developed, specific tasks still require customized modifications or adjustments.

On the other hand, large models enhance document layout analysis by correlating textual and visual features through pre-training. LayoutLM [9-11] introduced a diverse set of pre-training objectives, achieving outstanding performance in various document-related tasks. DiT [8] further strengthened its capabilities through self-supervised pre-training on large document datasets. Similarly, VGT [32] employed a grid-based textual encoding strategy to effectively extract text features, improving the accuracy of document understanding. LayoutReader [33] improved reading order detection by leveraging pretrained models, while DocParser [34] and DSG system [35] offered innovative solutions for parsing complex document structures. Additionally, RoDLA [36] established a key benchmark for assessing the robustness of document layout analysis models. These studies collectively demonstrate the potential of multimodal techniques in advancing layout analysis and provide valuable insights for future research directions and applications. However, these works are designed for general document analysis, which are not as accurate as expected for electronic component documents. Thus, specific analyzer should be developed.

## III. DATA ANALYSIS

A datasheet is a technical document that provides engineers with detailed information about the electronic components. This type of document details the electrical specifications of the device, such as the voltage and current tolerance range, as well as performance parameters, including operating temperature range, response time and gain. It also contains the physical packaging details of the device, which is crucial for circuit design and PCB layout. The datasheet guides engineers on how to integrate devices into their designs by providing typical application examples. In addition, it emphasizes the absolute maximum rating of the device to ensure the safe use of the device without exceeding these limits. Characteristic curves and diagrams help engineers understand how devices perform under different conditions, while test conditions ensure accuracy and repeatability of data. Reliability data, safety certification and environmental suitability information are also important elements of the data book to help engineers assess the long-term stability and compliance of devices. Fig. 1 shows the examples of common datasheets.

From the Fig. 1, we can find that the content of the datasheet is diverse, and often does not contain boundaries, it is difficult to cut and classify directly through the boundary line, in addition, there are many kinds of pictures, the content and function between various pictures are not the same, if no further distinction is made, it is still not conducive to engineers for rapid understanding and analysis. Moreover, the correlation between the picture and the caption is poor, and it is difficult to connect them using conventional machine learning methods.

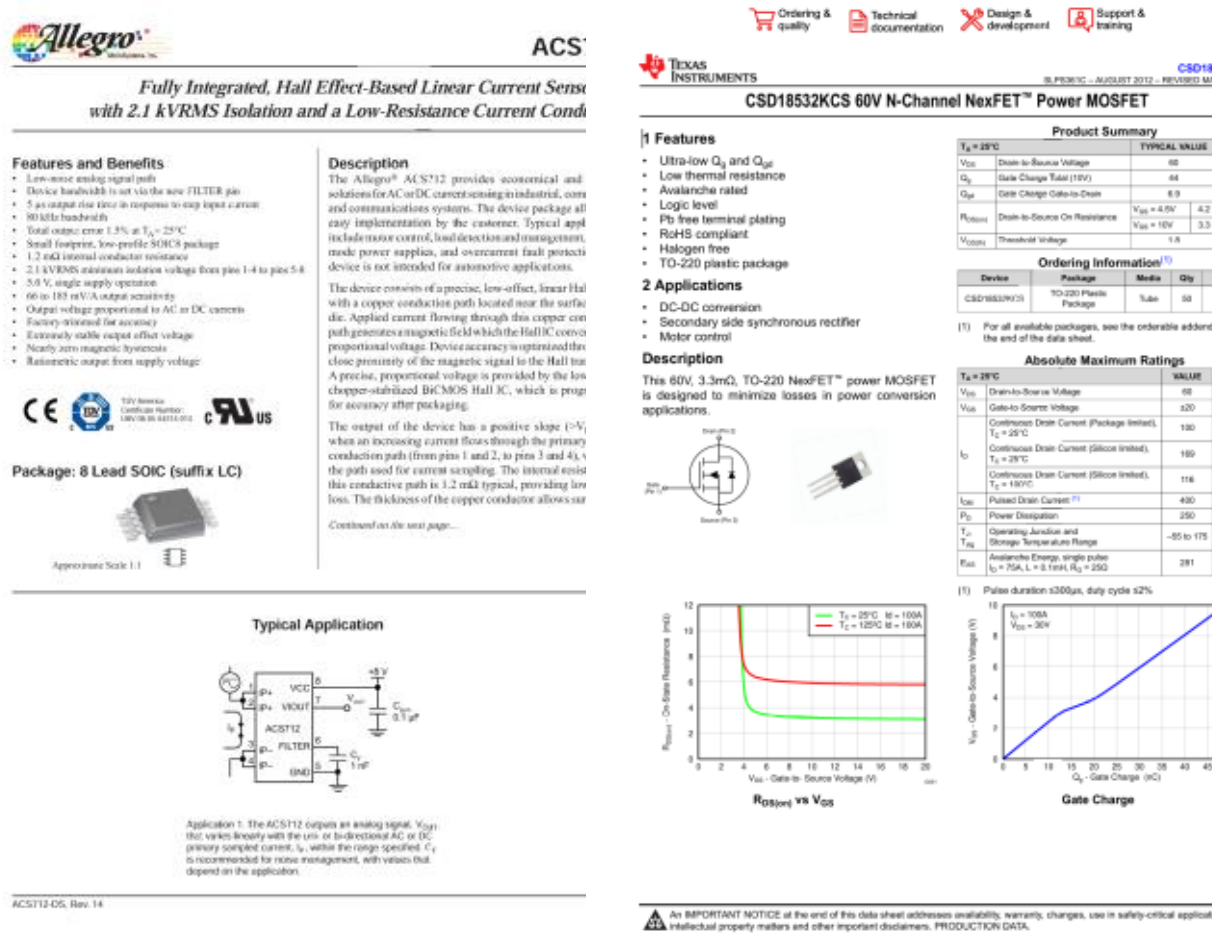

(a) (b)

Fig. 1. Example of a typical electronic component datasheet.

Compared to the existing document dataset, which classifies document layouts into only five categories: Text, Title, List, Table, and Figure, this paper categorizes the content of datasheets into 21 classes, including Text, Title, List, Table, functional block diagrams, flowcharts, characteristic curve diagrams, timing diagrams, etc. The dataset is tailored to the documentation of electronic component usage and characteristics, further refines the Figure category into a more detailed classification based on the role of the images. The typical example of each type is listed in Table I.

Additionally, the captions of images and tables are also an important part of obtaining information and should typically be analyzed together with the images or tables. However, in the traditional document dataset, they are categorized as text, which neglects its uniqueness. This paper separately labels the captions as a new category and save their content and position.

Finally, these datasheets are labeled referring to the COCO dataset format. It is important to note that the location information of the elements includes the top-left pixel coordinates and the width and height of the elements, expressed in an array: $[x_1, y_1, w, h]$. Having understood the characteristics and challenges of the electronic device document dataset, the next section details the EDocNet network and its training process.

## IV. METHODOLOGY

In this paper, a lightweight convolutional neural network composed of multiple depth separable layers is proposed to analyze electronic device documents and named as EDocNet network. As the training process shown in Fig. 2, the network is trained by focus and global knowledge distillation method, which significantly improves the prediction accuracy and reduces the prediction time.

### *C. The Network Structure of Backbone*

The model's backbone structure is presented in Fig. 3. Common convolutional layers must take into account both spatial information and channel correlation, followed by nonlinear activation of the output. Depthwise separable convolution, on the other hand, first conducts channel convolution, separates ordinary convolution in the spatial dimension, expands the network width, enriches the extracted features, and then performs point-by-point convolution. The impact of this decomposition is a significant reduction in the computational load and the model size [37].

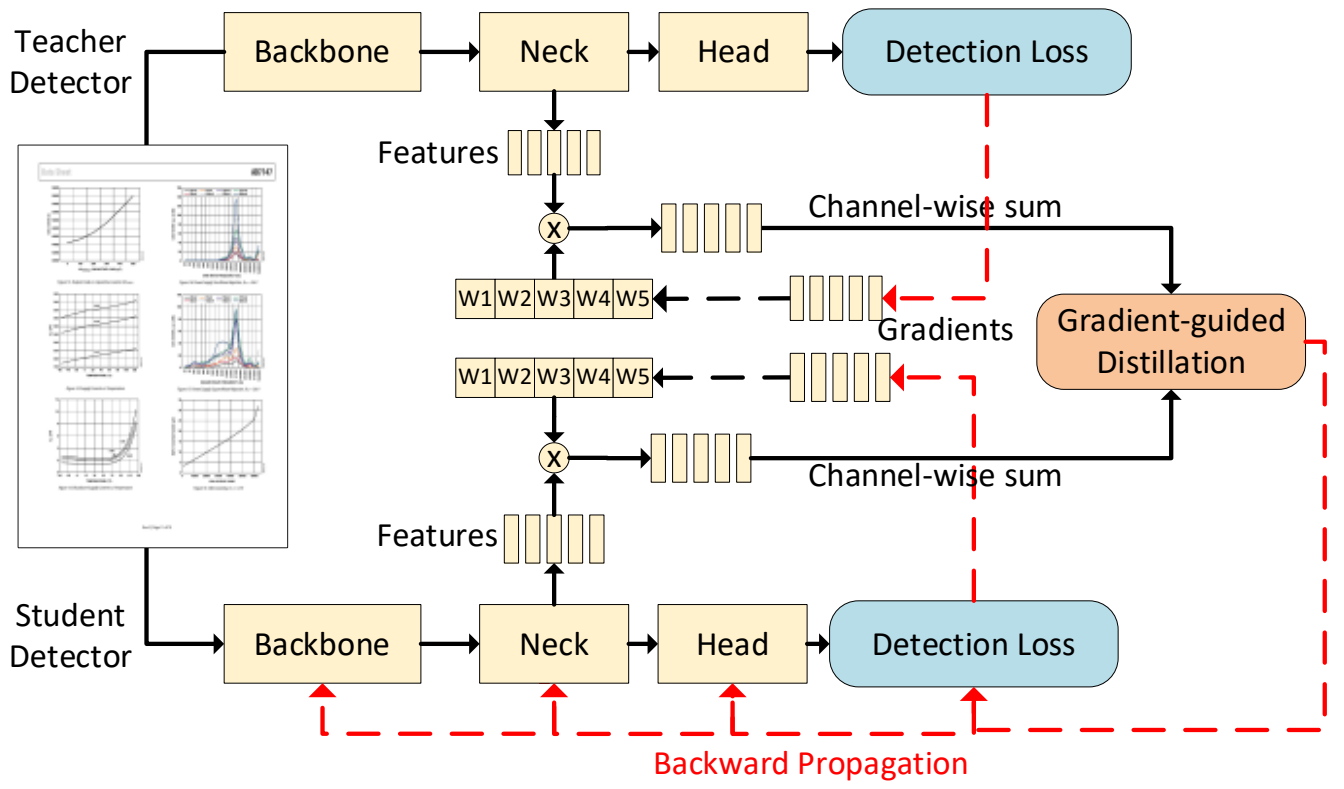

Fig. 2. The training flowchart of the proposed network using gradient knowledge distillation.

- **Depthwise separable convolution**

Depthwise separable convolution is a two-step process for convolving data. First, it performs depthwise convolution, which convolves each input channel separately. This means it focuses on one channel at a time, extracting spatial features unique to that channel. Then, it combines these results using pointwise convolution with 1x1 kernels. Depthwise convolution is responsible for extracting spatial information, while pointwise convolution focuses on extracting channel-related features, working together to provide an efficient and effective way to process and analyze data. This combination significantly reduces the number of calculations required compared to standard convolution while still effectively capturing important data features.

In the depthwise convolution operation shown in Fig. 3, $M$ convolution kernels of size $L{\times}L$ are used per convolution calculation, typically resulting in an output of 1. Pointwise convolution uses $M$ convolution kernels of $1{\times}1$ per operation. Evidently, depthwise and pointwise convolutions can be concatenated to form a standard convolution with a kernel size of $L{\times}L$ and $M$ channels. The number of parameters involved in the Depthwise separable convolution calculation of the combination of Depthwise convolution and Pointwise convolution are shown as follows:

$$\Delta_{DC,PC} = L \times L \times 1 \times M + 1 \times 1 \times M \times N \qquad (1)$$

$$R = \frac{\Delta_{DC,PC}}{\Delta_{CNN}} = \frac{L\times L\times 1\times M + 1\times 1\times M\times N}{L\times L\times M\times N} = \frac{1}{L^2} + \frac{1}{N} \qquad (2)$$

Using Depthwise separable convolution instead of standard convolution can reduce the time cost of convolution calculation.

- **H-Swish Activation function**

The commonly used activation functions are ReLU, Sigmoid. In this paper, a relatively novel activation function, H-Swish [40], is used. It is an improvement on the Swish activation

Table I The Classification Catalogues of the Datasheet Contents.

| Description | Item | Description | Item |
|---|---|---|---|
| **Functional Block Diagram** Display the functional components of electronic devices and their relationships, used to illustrate the overall architecture and division of functional modules. | 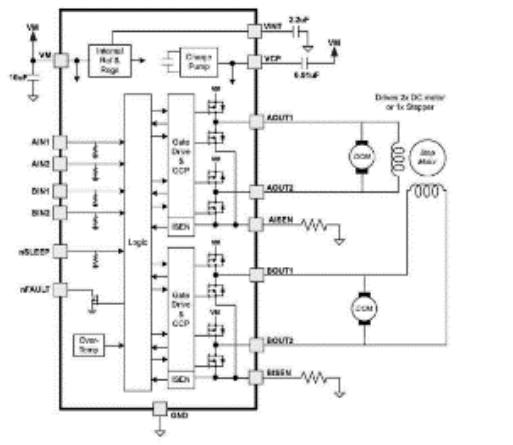 | **3D Schematic Diagram** Present the appearance or internal structure of electronic devices in three-dimensional graphics, providing more intuitive information about device shape and structure. | 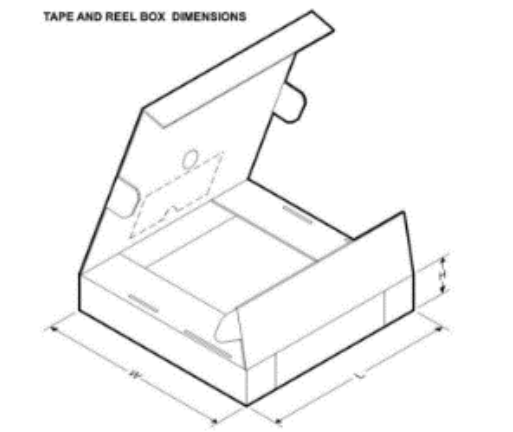 |
| **Flowchart** Illustrate the processes, operational sequences, or steps of algorithms in electronic devices or systems, used to explain workflow or control logic. | 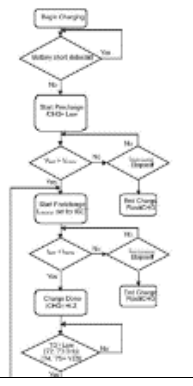 | **Pin Name Diagram** Display the numbering and corresponding functions of electronic device pins, assisting users in correctly connecting devices. | 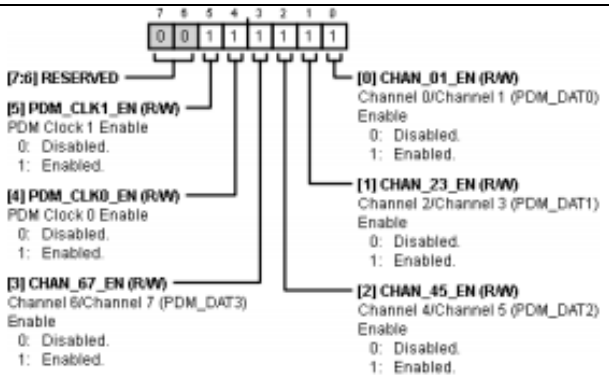 |
| **Characteristic Curve Diagram** Show how specific parameters of electronic devices change with inputs or other conditions, used to describe device performance and characteristics. | 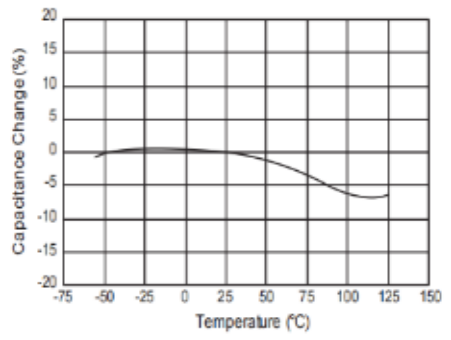 | **Marking Diagram** Add labels to various parts or functions of electronic devices to enhance readability and understanding. | 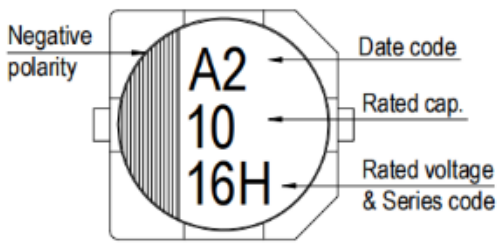  |
| **Timing Diagram** Describe the temporal sequence and relative timing relationships of different signals in electronic devices or systems, used to illustrate timing logic and requirements. | 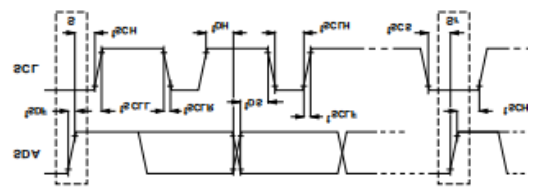 | **Appearance Diagram** Show photographs or images of the actual appearance of electronic devices, assisting in identification and visual inspection. | 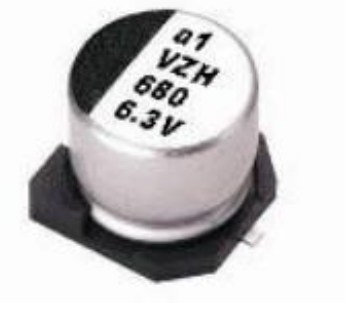 |
| **Circuit Diagram** Show the internal circuit connections and component layout of electronic devices, used to explain circuit design and connection methods. | 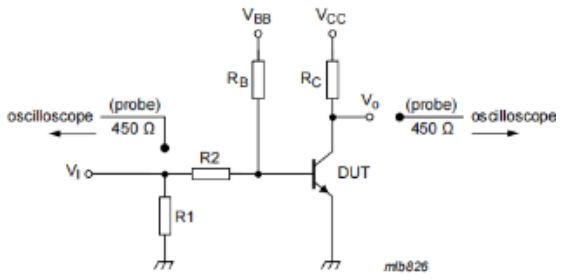 | **Functional Register Diagram** Describe the functions and bit fields of internal registers in electronic devices, explaining register configuration and operation methods. | 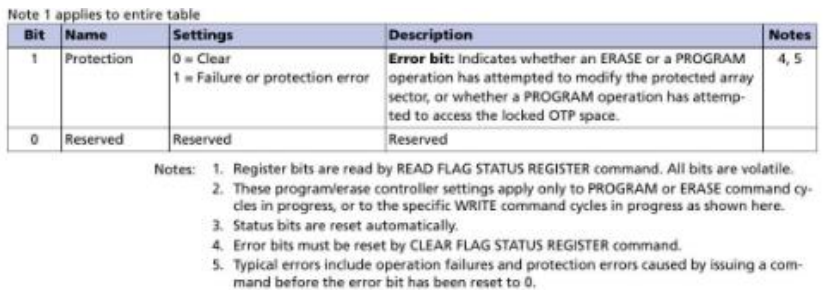 |
| **Pin Diagram** Display the layout of pins and their functions on electronic devices, used to explain how devices connect with external equipment. | 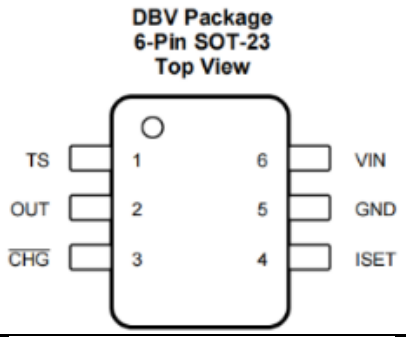  | **Layout Diagram** Display the layout and connection methods of internal components in electronic devices, explaining the layout of circuit boards and wiring. | 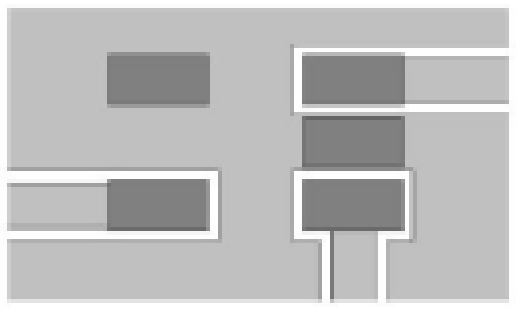 |
| **Engineering Drawing** Show the physical dimensions and external structure of electronic devices, used to describe size specifications and mounting requirements. | 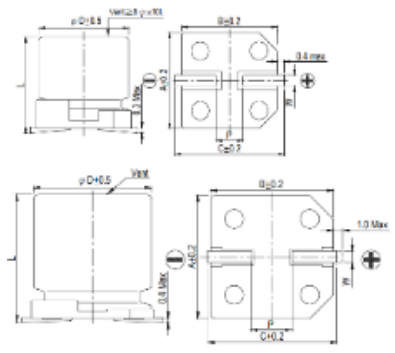 | **Data Structure Diagram** Describe the organization and relationships of internal data in electronic devices, explaining data structure and access methods. | 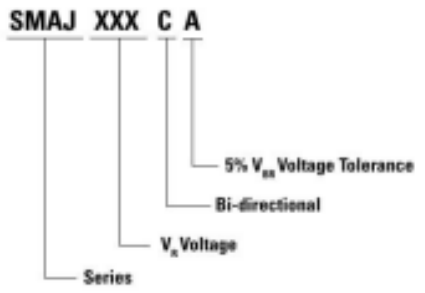  |
| **Sampling Diagram** Display the sampled data of electronic device output signals, used to describe signal waveforms and signal quality. | 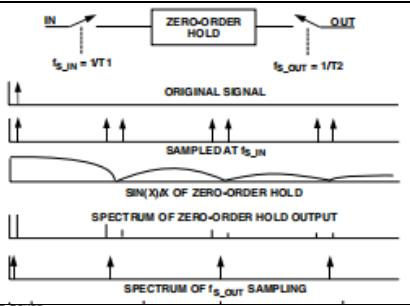  | **Text** Use text to describe the key information of the device. | 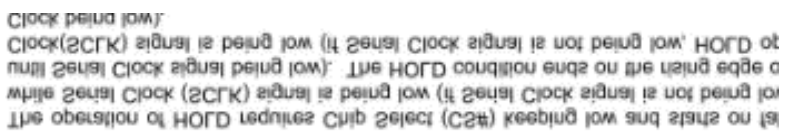 |
| **Table** Used to clearly display and compare key device parameters and characteristics. | 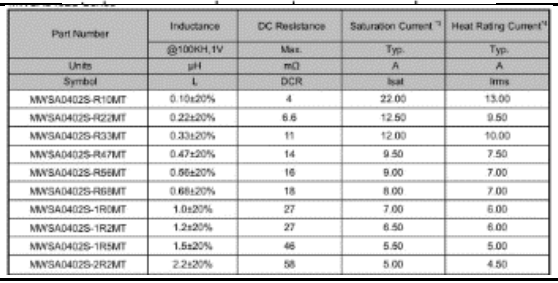 | **List** Provide ordered or unordered items, organize and present information in a concise and clear way. | 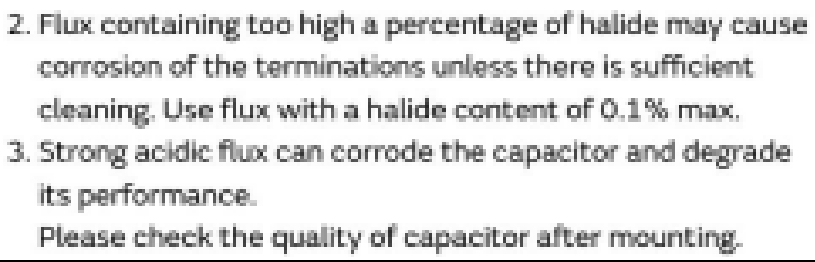  |
| **Title** Summarize and highlight the core content of a document or a specific section of a document. | 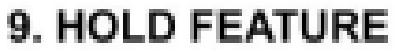  | **Caption** Provide instructions for visual elements to help readers understand its context. | 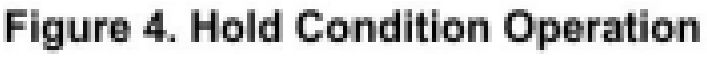  |
| **Other** Some of the fewer types of images. | | | |

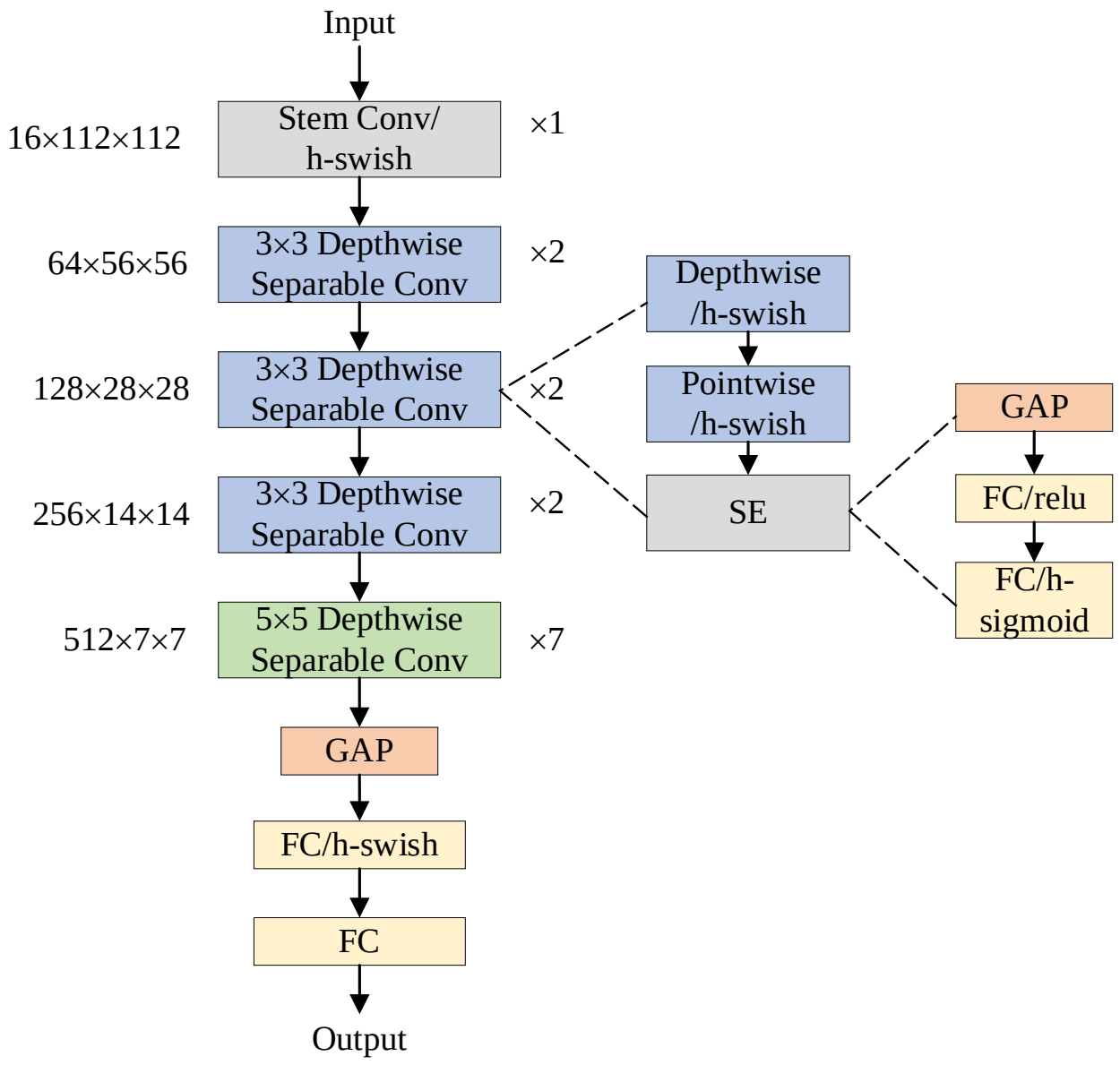

Fig. 3. The structure of backbone network of EDocNet.

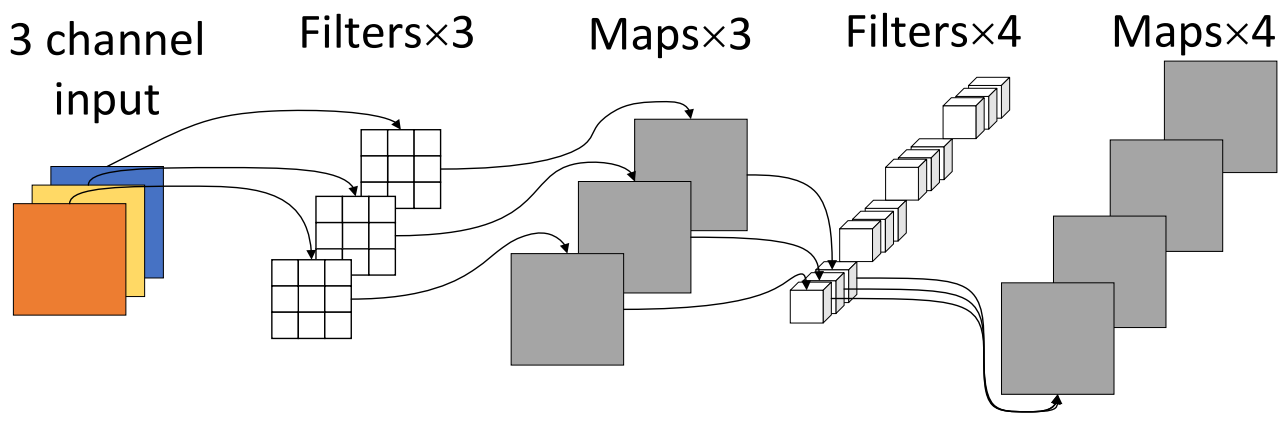

Fig. 4. The schematic of depthwise separable convolution.

function. The formulas are as follows:

$$H_{Swish(x)} = x \times \min(\max(0, x+3), 6)/6 \tag{3}$$

H-swish replaces the computationally expensive sigmoid function, reducing computational load and enabling faster computation. Unlike traditional sigmoid or tanh functions, H-swish mitigates the vanishing gradient problem, facilitating the training of deeper networks by maintaining healthier gradients. It also offers a smooth transition between nonlinear and linear regions, enhancing the model's ability to learn complex functions. Furthermore, H-swish's non-monotonic behavior for negative inputs allows the model to represent a broader range of functions.

- **The Squeeze-and-Excitation (SE) module**

The SE module is used to enhance the performance of convolutional neural networks. The SE module can be seen as a form of attention mechanism that allows the network to automatically learn the importance of different feature channels and adjust the contribution of each channel in the feature map accordingly. The SE module mainly includes three steps:

**Squeeze**: In this step, global average pooling (GAP) compresses the spatial dimensions (height and width) of each channel in the feature map into a single value, creating a channel descriptor. This allows each channel to obtain a global receptive field and capture global context information.

**Excitation**: After obtaining the channel descriptor, the importance of each channel is evaluated through two fully connected (FC) layers. The first FC layer reduces the number of channels and introduces non-linearity through the ReLU activation function. The second FC layer restores the number of channels and generates weights for each channel using the Sigmoid activation function.

**Scale**: Finally, the weights obtained from the excitation step are used to scale each channel of the original feature map. This is done by multiplying the learned channel weights with the original feature map on a per-channel basis, completing the feature recalibration.

### *D. Model Training Using Knowledge Distillation*

For the model training, Feature Gradient Distillation (FGD) is used, which aimed at efficiently transferring the deep feature and gradient information of a complex teacher model to a more concise student model. This approach enables the student model to inherit valuable knowledge and characteristics from the teacher model, facilitating its performance improvement [38, 39].

- **Focal Distillation**

Due to the imbalance between the front background, focal distillation is employed to classify the images and guide students to focus on key pixels and channels. The binary mask shown in Equ. (4) is used to separate the foreground information and background information of the image.

$$M_{i,j} = \begin{cases} 1, & if\ (i,j) \in r \\ 0, & Otherwise \end{cases} \tag{4}$$

where $r$ is the ground truth (also refers to GT bbox), and $i$ and $j$ represent the horizontal and vertical coordinates of the feature map respectively. If the point of the feature map falls within the bbox, the point is 1; otherwise, it is 0.

Since larger targets occupy a significant number of pixels, they tend to contribute disproportionately to the overall loss, thereby overshadowing the distillation process for smaller targets. Meanwhile, in different images, the ratio of the foreground to the background varies substantially. To address these issues and balance the losses of the varied elements, a scale mask $S$ is set:

$$S_{i,j} = \begin{cases} \frac{1}{H_r W_r}, & if\ (i,j) \in r \\ \frac{1}{N_{bg}}, & Otherwise \end{cases} \tag{5}$$

$$N_{bg} = \sum_{i=1}^{H} \sum_{j=1}^{W} (1 - M_{i,j}) \tag{6}$$

where $H_r$ and $W_r$ indicate the height and width of the GT bbox. If a pixel belongs to a different target, the smallest box is selected as GT bbox.

The absolute average of different pixels and different channels are obtained as follows:

$$C^S(F) = \frac{1}{C} \cdot \sum_{c=1}^{C} |F_C| \tag{7}$$

$$G^C(F) = \frac{1}{HW} \cdot \sum_{i=1}^{H} \sum_{j=1}^{W} |F_{i,j}| \tag{8}$$

where $H$, $W$, and $C$ represent the height, width, and channel of the feature. $G^S$ and $G^C$ are spatial and channel attention diagrams. the capital $S$ and $C$ in superscript represents spatial and channel attention respectively. $F$ is the input feature. $F_C$ indicates the feature of channel c. $F_{i,j}$ represents the feature of the pixel in row $i$ and column $j$. The spatial mask $A^S$ and channel attention mask $A^C$ are used to distinguish key pixels and channels and can then be described as:

$$A^S(F) = H \cdot W \cdot softmax(G^S(F)/T) \tag{9}$$

$$A^C(F) = C \cdot softmax(G^C(F)/T) \tag{10}$$

where $T$ is the temperature overparameter used to adjust the distribution.

There are significant differences between students' and teachers' masks. During the training process, the teacher's masks are used to guide the students'. Based on the above results, the feature loss function of the focal loss is calculated:

$$L_{fea} = \alpha \sum_{k=1}^{C} \sum_{i=1}^{H} \sum_{j=1}^{w} M_{i,j} S_{i,j} A_{i,j}^{S} A_{k}^{C} (F_{k,i,j}^{T} - f(F_{k,i,j}^{St}))^2 + \beta \sum_{k=1}^{C} \sum_{i=1}^{H} \sum_{j=1}^{w} (1 - M_{i,j}) S_{i,j} A_{i,j}^{S} A_{k}^{C} (F_{k,i,j}^{T} - f(F_{k,i,j}^{St}))^2 \quad (11)$$

where $F^T$ and $F^{St}$ represent the feature graphs of the teacher detector and the student detector respectively. The superscript $St$ and $T$ represents student and teacher respectively. $\alpha$ and $\beta$ are hyperparameters that balance the loss between foreground and background. $A^S$ and $A^C$ represent the spatial attention mask and channel attention mask of the teacher detector, respectively. $M_{i,j}$ represents the binary mask of the pixel.

In addition, the paper uses the attention Loss Lat to force the student detector to mimic the spatial and channel attention mask of the teacher detector:

$$L_{at} = \gamma(l(A_T^S, A_{St}^S) + l(A_T^C, A_{St}^C)) \quad (12)$$

where $l()$ represents L1 loss, and $\gamma$ is a hyperparameter of equilibrium loss.

The final key losses are as follows:

$$L_{focal} = L_{fea} + L_{at} \quad (13)$$

- **Global Distillation**

In focus distillation, images are separated by local distillation and students are forced to focus their attention on key parts. This distillation severs the relationship between foreground and background. Therefore, this paper also proposes global extraction, which aims to extract the global relationship between different pixels from the feature map to make up for the global information loss of the key distillation.

Global distillation is used to extract the relationship between different pixels and enhance the relationship between the front and background. The global loss $L_{gloabl}$ is as follows:

$$L_{global} = \lambda \cdot \sum (R(F^T) - R(F^{St}))^2 \quad (14)$$

$$R(F) = F + W_{v2}(ReLU(LN(W_{v1}(\sum_{j=1}^{N_p} \frac{e^{W_k F_j}}{\sum_{m=1}^{N_p} e^{W_k F_M}} F_j)))) \quad (15)$$

where, the inputs are the neck layer of the teacher network and the student network respectively, $W_k$, $W_{v1}$ and $W_{v2}$ represent the convolutional layer, $LN$ represents the layer normalization, $N_p$ represents the number of pixels in the feature, and $\lambda$ is the hyperparameter of the balance loss.

The final loss includes two parts: original training loss and distillation loss, and distillation loss includes local loss and global loss, as follows:

$$L = L_{original} + L_{focal} + L_{global} \quad (16)$$

## V. EXPERIMENTAL CONTENT

To validate the proposed method, the method is evaluated on the datasheet dataset which contains 8677 images from the documentation of 500 electronic devices, including capacitor, connector, detector, diode, inductor, resistor, etc. The performance indexes of the algorithm used in this paper are Average Recall (AR) and Average Precision (AP).

AR is a measure of the average recall rate of a model under different confidence thresholds. The recall rate is the percentage of all samples that are actually positive that are correctly predicted by the model to be positive. The formula for AR is as follows:

$$AR = \frac{TP}{TP+FN} \quad (17)$$

where $TP$ is the number of samples correctly identified as positive. $FN$ is the number of samples that are positive but misidentified as negative.

AP is a measure of the average accuracy of a model at different thresholds. Accuracy refers to the proportion of all samples predicted by the model to be positive that are actually positive. The AP is expressed as follows：

$$AP = \frac{TP}{TP+FP} \quad (18)$$

where $FP$ is the number of samples that are negative but misidentified as positive.

IoU is a measure of the accuracy of detecting objects in a particular dataset. That is, the overlap rate between the generated candidate bound and the original labeled box (ground truth bound). The ratio of their intersection to their union. The best case is a perfect overlap, that is, a ratio of 1.

The Area in the evaluation indicator usually refers to the area of the detection box or target area in the target detection result, and is divided into different categories (such as Small, Medium, and Large) according to the size of the target area, so as to more accurately evaluate the performance of the detector on the target of different scales.

### *E. Training result*

The experimental results of the model are shown in Table II and III. MaxDets limits the maximum number of objects detected in each image. Since the number of targets in document parsing is generally large, we will only show and analyze cases where MaxDets=100. First, the AP of the model in all target categories, all sizes and IoU threshold ranges (0.50:0.95) is 0.765, indicating that the model can maintain high detection accuracy under the comprehensive consideration of different IoU thresholds. Under a certain IoU threshold, when IoU=0.50, the average accuracy of the model is 0.711, and when IoU=0.75, the average accuracy of the model is 0.686, which further verifies the stability and reliability of the model in a wide range of IoU thresholds. For targets of different sizes, the average accuracy of the model is 0.730 for medium targets and 0.666 for large targets. However, for small targets, the model cannot provide specific performance values, possibly because there are fewer small targets in document layout analysis.

TABLE II
AVERAGE PRECISION RESULTS

| maxDets | IoU | Area | Average Precision |
|---|---|---|---|
| 100 | 0.50：0.95 | All | 0.765 |
| 100 | 0.50 | All | 0.711 |
| 100 | 0.75 | All | 0.686 |
| 100 | 0.50：0.95 | Medium | 0.73 |
| 100 | 0.50：0.95 | Large | 0.666 |
| 100 | 0.50：0.95 | Small | -1.000 |

TABLE III
AVERAGE RECALL RESULTS

| maxDets | IoU | Area | Average Precision |
|---|---|---|---|
| 100 | 0.50：0.95 | All | 0.934 |
| 100 | 0.50 | All | 0.915 |
| 100 | 0.75 | All | 0.916 |
| 100 | 0.50：0.95 | Medium | 0.784 |
| 100 | 0.50：0.95 | Large | 0.916 |
| 100 | 0.50：0.95 | Small | 0.655 |

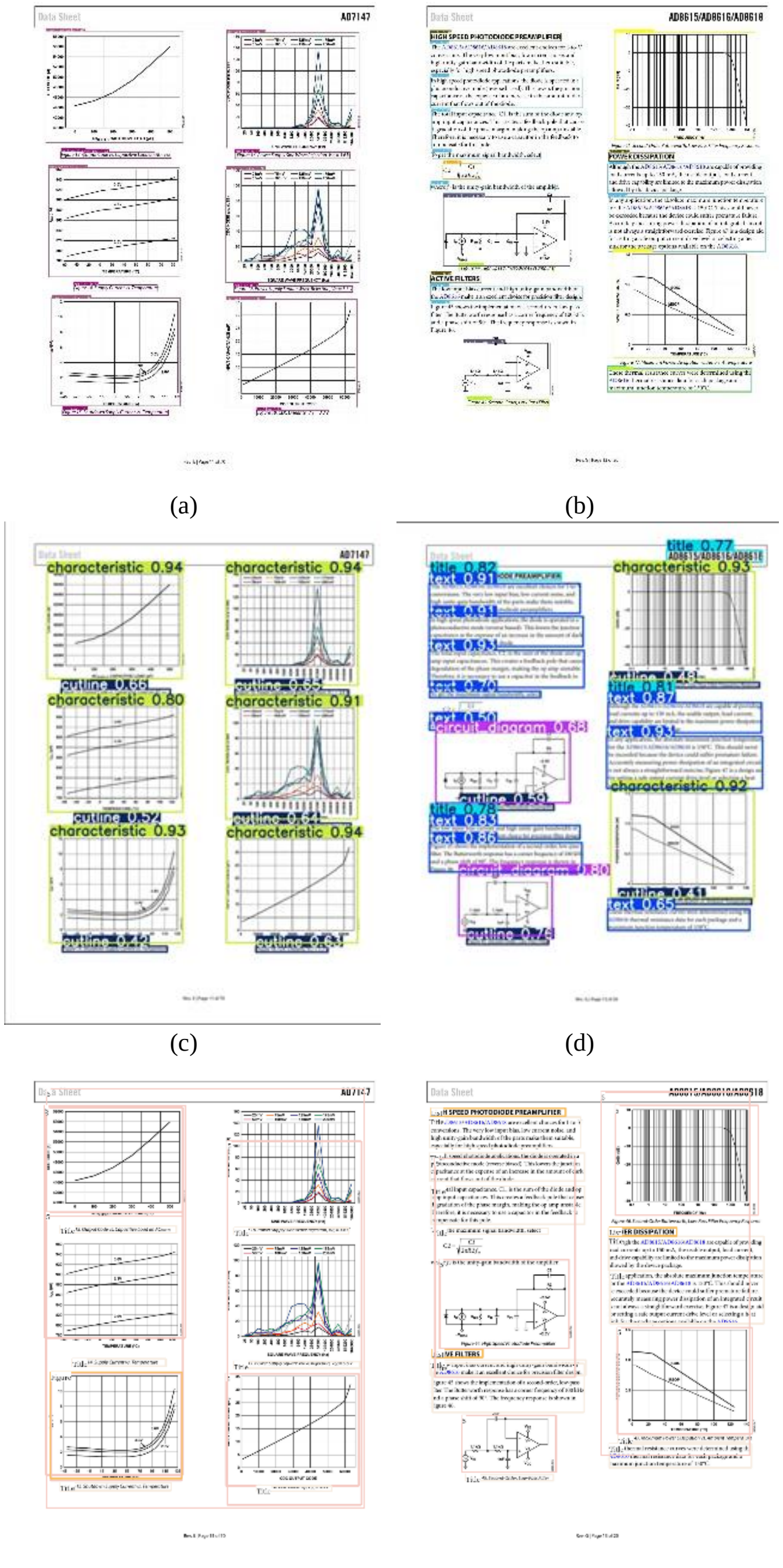

Fig. 5. Layout analysis comparison. (a), (b) EDocNet. (c), (d) YOLOv8. (d), (e) faster R-CNN model .

From the perspective of AR, when the maximum number of detections is 100, the average recall rate within each IoU threshold range (0.50:0.95) reaches 0.934, indicating that the model can effectively identify most targets. When the maximum detection number is 1 and 10, the average recall rate of the model is 0.655 and 0.915, respectively, indicating that the recall ability of the model still maintains a high level under different restrictive conditions. Especially for the large target, the average recall rate of the model also reached 0.916, while the average recall rate of the medium target was 0.784, which once again proved the excellent performance of the model on the large target.

### *F. Compared with Target Detection Models*

Therefore, this paper compares the state-of-art (SOTA) detection models for comparative evaluation including Faster R-CNN, Mask R-CNN, YOLO, and recently proposed DocLayout-yolo and D-FINE. Faster R-CNN employs a Region Proposal Network (RPN) to expeditiously generate region proposals, which are subsequently subjected to classification and bounding box refinement via a convolutional neural network. Mask R-CNN is an extension of the Faster R-CNN object detection framework that incorporates an additional branch for instance segmentation. It allows for the simultaneous identification and segmentation of multiple objects within an image. YOLO predicts both bounding boxes and class probabilities for a given image in one pass through the network and is known for its speed and simplicity. The yolo model for comparison in this work is Yolov8 and DocLayout-YOLO [41]. DocLayout-YOLO combines YOLO's efficient inspection capabilities with the needs of document analysis to provide a more accurate and efficient document structure inspection solution through customized adjustments to document layout. D-FINE [42] redefines box regression with FDR (fine-grained distribution optimization) and GO-LSD (global optimal localization self-distillation). It simplifies optimization, models uncertainty, and shares localization knowledge, achieving high performance in real-time object detection. The comparative experimental results are shown in the Table IV.

TABLE IV
PERFORMANCE OF DIFFERENT MODELS WHEN IOU=0.50:0.95,AREA=ALL

| Model name | AP | AR | Time |
|---|---|---|---|
| Faster R-CNN | 0.612 | 0.816 | 12.3s |
| Mask R-CNN | 0.634 | 0.852 | 15.6s |
| Yolov8 | 0.684 | 0.675 | 1.9s |
| DocLayout-yolo | 0.653 | 0.864 | 1.5s |
| D-FINE | 0.682 | 0.921 | 2.6s |
| **EDocNet** | **0.765** | **0.934** | **0.236s** |

The comparative analysis presented in the table demonstrates the superior performance of the EDocNet model in the task of document layout analysis. EDocNet exhibited the highest average recall (AR) of 0.916, surpassing the performance of Faster R-CNN, Mask R-CNN, and YOLOv8 models. Concurrently, it achieved a commendable AP of 0.765. Notably, the EDocNet model demonstrated exceptional processing speed, with an inference time of merely 0.236 seconds per image, which is significantly faster than the other models under consideration. These findings underscore the EDocNet's capability to maintain high detection accuracy while offering expedited processing, positioning it as a formidable contender for applications in datasheet analysis.

The comparison of prediction effects of EDocNet, YOLOv8 and Faster R-CNN is shown in Fig. 5. It can be seen from the figure that both EDocNet and YOLOv8 have higher prediction effect than Faster R-CNN. Faster R-CNN has produced many identification results that should not exist. It also classifies images correctly when making predictions.

### *G. Compared with Large Models*

The development of large models is particularly rapid in the field of document layout analysis, so we also compare several large models in the field. LayoutLMv3 is a multimodal pre-trained model by Microsoft that handles text and images in documents, suitable for various document AI tasks. DiT is a self-supervised pre-trained model for document images, addressing the issue of insufficient labeled data. PicoDet is a lightweight, fast object detection model designed for efficient real-time inference on resource-constrained devices. RT-DETR-H combines efficient convolutional neural networks and Transformer architecture to optimize inference speed and accuracy, and is designed for real-time object detection tasks.

TABLE V
PERFORMANCE OF DIFFERENT MODELS WHEN IOU=0.50:0.95,AREA=ALL

| Model name | AP | AR | Time |
|---|---|---|---|
| LayoutLMv3 | 0.563 | **0.944** | 0.3s |
| Dit | 0.507 | 0.908 | 0.6s |
| PicoDet | 0.459 | 0.738 | 0.568s |
| RT-DETR-H | 0.501 | 0.771 | 0.621s |
| EDocNet without FGD | 0.665 | 0.916 | 0.346s |
| **EDocNet** | **0.765** | **0.934** | **0.236s** |

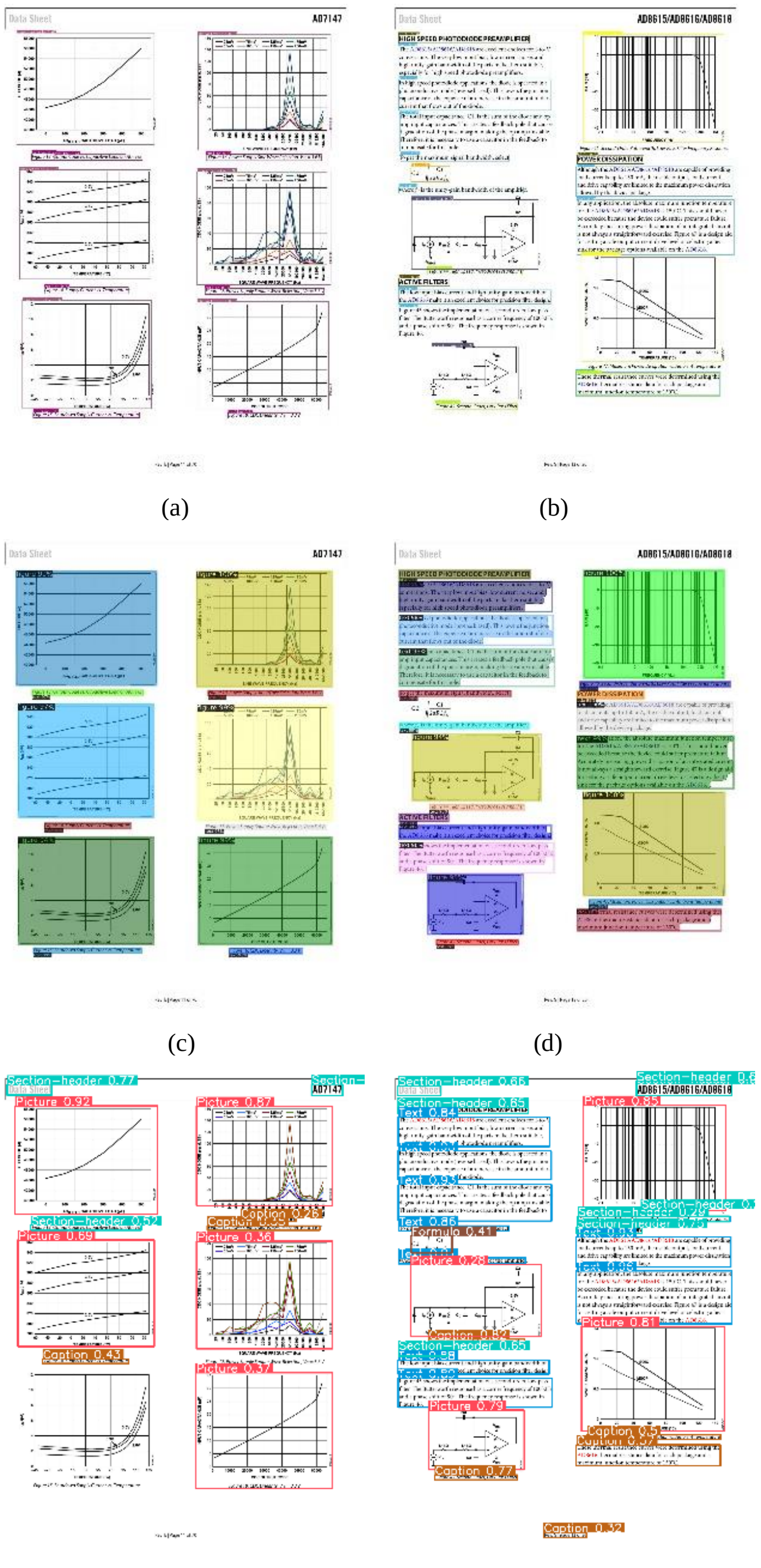

(a) (b) (c) (d) (e) (f)

Fig. 6. Layout analysis results comparison. (a), (b) EDocNet. (c), (d) LayoutLMV3. (d), (e) Dit.

The comparison with the five large models is shown in Table V and Fig. 6. As can be seen from the figure, compared with other large models, this model can not only accurately analyze document layout, but also further classify pictures according to the characteristics of electronic device documents. For example, in Fig.6(a), compared with Fig.6(c) and (e), the category of figure predicted by the other two large models is further subdivided into characteristic curve pictures, and the cutline part and the text part are distinguished.

It can be seen from the table that EDocNet has the best performance in accuracy (AP), recall rate (AR) and inference time, with the accuracy up to 0.765, recall rate of 0.934, and inference time of only 0.236 seconds, which is suitable for scenarios requiring high precision and fast response. In contrast, LayoutLMv3 performs well on recall (0.944), but has lower accuracy (0.563) and shorter reasoning time (0.3 seconds). PicoDet and RT-Dedr-H are medium in accuracy and recall, and have moderate reasoning times. The Dit's overall performance was weaker, with lower accuracy and recall rates than other models.

## VI. CONCLUSION

In this paper, we have proposed the EDocNet model which employs the focal and global knowledge distillation training approach for the refinement of specific target categories in the layout analysis of electronic device documents. Through experimental verification, this model has demonstrated a powerful detection ability and robustness in the layout analysis of electronic device documents, performing excellently in multiple evaluation metrics and possessing outstanding detection capabilities.

Compared with traditional object detection models such as Faster R-CNN, Mask R-CNN, and YOLOv8, the EDocNet model has achieved certain improvements in terms of average precision and average recall, and the training and prediction time per image has been significantly reduced. This makes it particularly suitable for the analysis of multi-page electronic device documents. In contrast to existing large models like LayoutLMv3 and DiT models, the EDocNet model divides the images into 16 categories and can accurately predict information of different categories. When delving into electronic device documents, these subdivided categories help the model analyze the performance of electronic devices more efficiently and provide engineers with more valuable information. Thus, it has significant advantages in the field of electronic device document analysis and is expected to be widely applied in relevant applications and promote the further development of this field.